# COMMUNIWAVE

*A Machine Learning Model for Quantifying the Degree of Temporary Informal Behavior in Urban Communities*

HONGYE YANG, SHIEN LIU
*Beijing Institute of Architectural Design Co., Ltd*
*yanghongye@biad.com.cn , hyang783@gatech.edu*
*Wuxi Taihu University*
*212017075@wxu.edu.cn*

AND

ZHIHAO XIE
*Wuxi Taihu University*
*212017074@wxu.edu.cn*

**Abstract.** For urban managers and designers, improving the functional attributes of urban communities to enhance territorial resilience in the face of complexity and uncertainty is crucial. Currently, community planning often follows a top-down approach and lacks effective metrics to quantify informal behaviors of residents, leading to frequent conflicts with original plans. This study introduces CommuniWave, a machine learning model designed to efficiently detect and quantify the Degree of Informal Behavior (DIB) in urban communities. The model integrates a Behavior Capture Net (BCN) based on mmaction2, a self-developed YOLOv10 model (YLX), and a Behavior Eval Model (BEM) using random forest. Ultimately, by generating DIB fluctuation charts from street videos, the model facilitates dynamic monitoring, supporting urban managers in making refined decisions to enhance the overall resilience of communities.

## 1. Introduction

Territorial resilience is an interdisciplinary concept of sustainable development(Brunetta et al., 2019), emphasizing the need for decision-makers to adopt flexible and proactive approaches in addressing risks and challenges within a given space. In urban communities, spontaneous bottom-up activities by residents represent uncertain factors. However, management approaches toward these activities often lack sustainable understanding. Previous studies have shown that a community management model combining bottom-up and top-down approaches helps reduce conflicts, thereby promoting sustainable territorial resilience(Semeraro et al., 2020).

Informal behavior in urban communities is an informal phenomenon whose research origins can be traced back to economist Keith Hart's discussion of the Informal Sector(Hart, 1985). In communities, the disconnect between government-led urban planning and the daily needs of residents is the main reason for the emergence of informal behaviors(Anon., 2018). These behaviors often manifest as spontaneous, context-specific responses, such as randomly appearing street stalls or spontaneously formed gathering places(Chase, Crawford and Kaliski, 2008). These informal behaviors of community residents often accompany rapid urban expansion and unbalanced land policies, reflecting issues of imbalance between infrastructure supply and demand(Frederic Deng and Huang, 2004).

Scholars generally believe that these spontaneous activities have positive significance for community life. Jane Jacobs, in The Death and Life of Great American Cities, emphasized that urban planning should create conditions for “casual public interactions” in community life(Jacobs, 1992); Jan Geh pointed out that the vitality of street spaces depends on social interactions in daily life(Gehl, 1971); Margaret Crawford argued that urban managers need to understand and accept the visual "disorder" brought about by the "counter-publics" actions of street vendors and the homeless(Crawford, 1995). Through residents' life experiences, these behaviors continuously reshape and redefine public spaces and domains, promoting the diversity and vitality of neighborhood activities(Mehta and Bosson, 2021).

In contemporary society, urban managers have begun to try to guide these bottom-up spontaneous behaviors within a compliant framework (Yao et al., 2024). In Beijing, China, dedicated community duty planners (CDPs) have been introduced to coordinate neighborhood issues(Zhou et al., 2023). This top-down communication public participation model has also been applied in the governance of urban communities in Nanjing, China(Cao, 2022). However, the evaluation of these models often relies on the experience of decision-makers and planners, lacking effective quantitative measures.

Although systematic methods such as Post-Occupancy Evaluation (POE)(Zimring and Reizenstein, 1980) and Public Space & Public Life

Survey (PSPL)(Lv and Ding, 2021) have been used to assess the use and satisfaction of urban spaces, these qualitative methods rely more on sociological field surveys, which are inefficient and have limited data accuracy.

Some emerging computational survey tools are helping designers assess and quantify human activity in cities. For example, semantic and spatiotemporal clustering methods analyze crowd movement patterns through smartphone Twitter check-in data(Steiger et al., 2015); residence location information provided by mobile data can compare the differences in activity range, anchor point quantity, and movement frequency among populations in different cities(Xu et al., 2018); by combining POI classification with automatic annotation algorithms, activity types can also be predicted(Furletti et al., 2013). However, these quantitative analysis methods are not practical for collecting data on informal behaviors in cities, as the behavior of the research subjects is diverse and the data sources are fragmented, making it difficult to fully reflect the overall informal behavior on the streets.

Recent developments in machine learning have enabled researchers to improve data collection efficiency through high-precision analysis of low-cost video and photo records. Hou and others used deep convolutional neural networks to analyze human behavior in videos and quantify the use of small public spaces(Hou et al., 2020); Li and others assessed the vitality of different streets by analyzing behaviors captured by surveillance cameras using deep learning models(Li, Yabuki and Fukuda, 2022). Although these previous studies effectively identified human activities and preferences, due to the complexity of informal behaviors in urban communities, scholars still need to develop more suitable classification models and workflows.

This paper introduces a machine learning-based model capable of rapidly identifying the Degree of Informal Behavior (DIB) in spaces and scoring them. Compared to traditional workflows, this approach enhances the research efficiency and precision management capabilities of urban designers, advancing the process of territorial resilience construction based on urban communities.

## 2.Methodology

The paper proposes a machine learning model named CommuniWave, which is divided into two parts. The first part is the behavior recognition model group (Behavior Capture Net, BCN), which uses mmaction2 developed on the PyTorch framework and a self-developed model YLX based on YOLOv10 to identify various community informal behaviors in input videos and record the original feature data of each behavior. The second part is the behavior evaluation model (Behavior Eval Model, BEM), a machine learning model based on random forests. By utilizing Principal Component Analysis (PCA) for optimization, it filters out the most contributive features while reducing dimensionality to minimize training times and prevent overfitting. Ultimately, CommuniWave is able to output a map of informal behavior changes on the input streets (see Figure 1）. For BEM, we replaced traditional linear models with machine learning models because they are better suited for the nonlinearity of complex data such as community informal behaviors(Mora-Garcia, Cespedes-Lopez and Perez-Sanchez, 2022).Additionally, compared to traditional questionnaire-based surveys, CommuniWave can greatly enhance research efficiency. Researchers can immediately analyze video data after collection, and the system can operate continuously, eliminating the fatigue and long working hours associated with manual labor. It also resolves the difficulties of distributing questionnaires effectively.

## 2.1 DATA COLLECTION

The identification and training process of the CommuniWave model relies on manually annotated datasets. First, when training the BEM, due to the varying attractiveness of different urban spaces to people, residents have different standards for determining informal behaviors on different street levels Therefore, we need to conduct a detailed analysis of various types of street spaces. We collected 180 video segments from three types of roads as the training dataset, including: 1. Streets that allow motor vehicles and non-motorized vehicles (First-class road, 60 segments); 2. Streets that do not allow motor vehicles (Second-class road, 60 segments); 3. Streets that only allow pedestrian traffic (Third-class road, 60 segments).

Due to limitations in current hardware and algorithms, no matter how large the training dataset provided, the model cannot entirely overcome geographical and cultural constraints. Moreover, large-scale models face significant computational demands during both the training and inference stages. After comparing the performance-to-cost ratios of different solutions, we opted for a smaller dataset and model. A smaller model offers lower inference costs and is easier to train. We also aim to provide a more flexible and customizable framework, allowing researchers to conduct tailored training based on the unique behaviors in their specific tasks. For YLX, we selected the highly efficient YOLOv10 framework and opened the training interface, enabling researchers to easily train their own YLX version by providing their own dataset. Researchers can also retrain the BEM to adapt to the scoring of informal behaviors specific to certain groups. Consequently, CommuniWave can better meet the needs of different community managers. To address privacy concerns related to video data collection in public spaces, we utilized de-identification methods (Cao et al., 2021) to prevent the invasion of personal privacy. This approach completely separates the association between personal identities and behavior information recognized by the model, ensuring that the model and researchers comply with relevant privacy regulations.

Second, when training the BCN, since the mmaction2 model can only recognize basic behaviors (such as walking, sitting, cycling, running, etc.), we trained the YLX model based on YOLOv10 to identify informal behaviors. We selected six common informal behaviors in Chinese communities (square dancing, gathering, street vending with and without equipment, three wheeled motorcycling, and chess playing) as training labels for the YLX model. The image data corresponding to these labels mainly came from Chinese social media platforms.

To accurately capture pedestrians' informal behaviors on the streets, all 180 video segments were filmed by cameras positioned 5-10 meters above the ground, located in the urban area of a medium-sized city in southern China. High-positioned cameras effectively avoid occlusion and crowd overlap issues in high pedestrian traffic situations and adapt to different site environments. The recorded videos were edited and scaled to the optimal size suitable for survey research and model training, with each video ultimately being 10 seconds long (see Figure 2）.

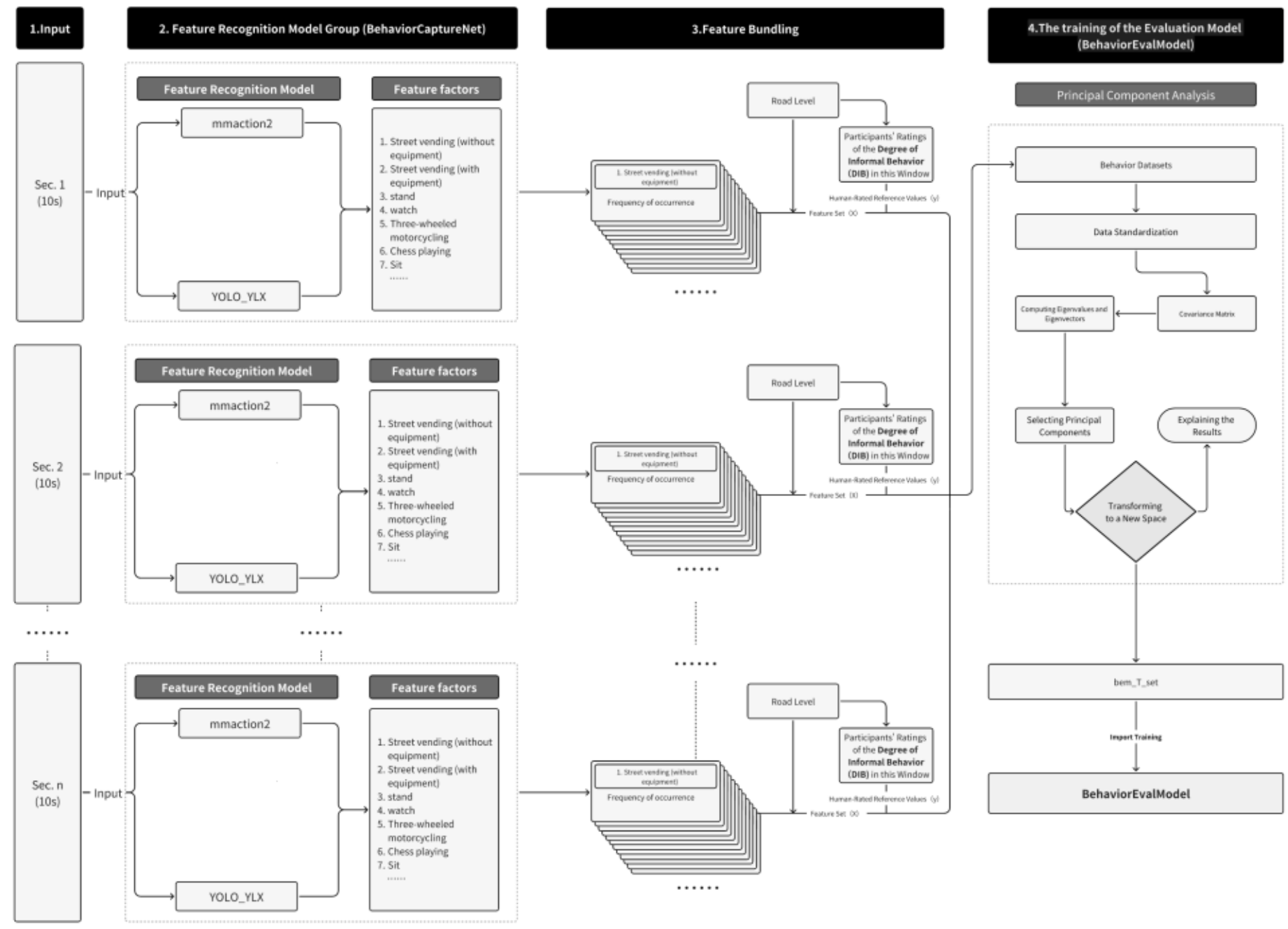

*Figure 1.* CommuniWave model training process.

## 2.2 PARTICIPANTS

This study strictly adhered to ethical standards and legal regulations, recruiting 10 volunteers (6 males and 4 females, with an average age of 31.0 ± 13.89 years). All volunteers signed written informed consent forms, allowing the anonymous use of their rating data. The volunteers were all residents of urban communities and were familiar with community life, ensuring that they could accurately assess the level of informal behavior in the video content. The survey included 180 video clips, and volunteers were asked to rate each clip on a scale of 1 to 5, with the scores representing: "1 - No temporary activities; 2 - Slightly some temporary activities; 3 - Moderate level of temporary activities; 4 - Significant temporary activities; 5 - Very intense temporary activities." To ensure objectivity in the ratings, the order of video playback was randomized, and no time limit was imposed on the volunteers for rating. To integrate all volunteer ratings while considering the sensitivity of minority opinions and reducing the complexity of model training, we used the mean value, rounded to two decimal places, as the informal behavior score (label y) for each video. To avoid introducing bias and subjectivity from volunteers into our predictive model, we adopted the Median Absolute Deviation (MAD) method (Xu et al., 2020) to filter out and eliminate extreme outliers with significant bias in volunteer ratings. Additionally, we compiled a diverse list of volunteers with varied backgrounds, ensuring that each set of scores is provided by volunteers from different educational levels, age groups, and genders. This approach helps to prevent the influence of bias and subjectivity from any particular demographic group.

## 2.3 BEHAVIOR CAPTURE NET（BCN）

The mmaction2 toolkit is an open-source package based on PyTorch, which supports a variety of video understanding models including action recognition, skeleton-based action recognition, spatiotemporal action detection, and temporal action localization. This toolkit was developed and released by open-mmlab on GitHub at mmaction2 GitHub repository. We utilized one of its pre-trained models for spatiotemporal action detection in videos, which not only identifies specific actions within the video but also accurately locates these actions along the timeline.

Furthermore, we have adopted the YOLOv10 model to train our YLX (Wang et al., 2024). This model features a one-to-one head design that eliminates the need for Non-Maximum Suppression (NMS) during inference, thereby reducing latency and enhancing efficiency. It is particularly suitable for informal behavior recognition tasks in complex scenes. To train the YLX model, we have curated a dataset comprising 30,000 images. Providing high-quality annotations for this dataset is of utmost importance and will be detailed in Section 2.6. During the inference phase, we combined the mmaction2 framework with the YLX model to achieve more efficient multimodal data processing and analysis.

## 2.4 FEATURE BUNDLING

After completing the inference process with BCN, we obtain three different datasets: Feature factors, the corresponding street levels for each video, and the volunteer score sheets. We need to integrate these three sets of data through a process called Feature Bundling. The initial Feature factors dataset is unordered, so we must standardize the format of the Feature factors for different videos, then match them one by one with the corresponding street levels and volunteer score sheets according to the video file number, combining them into a single, complete .csv file (bem_T_set).

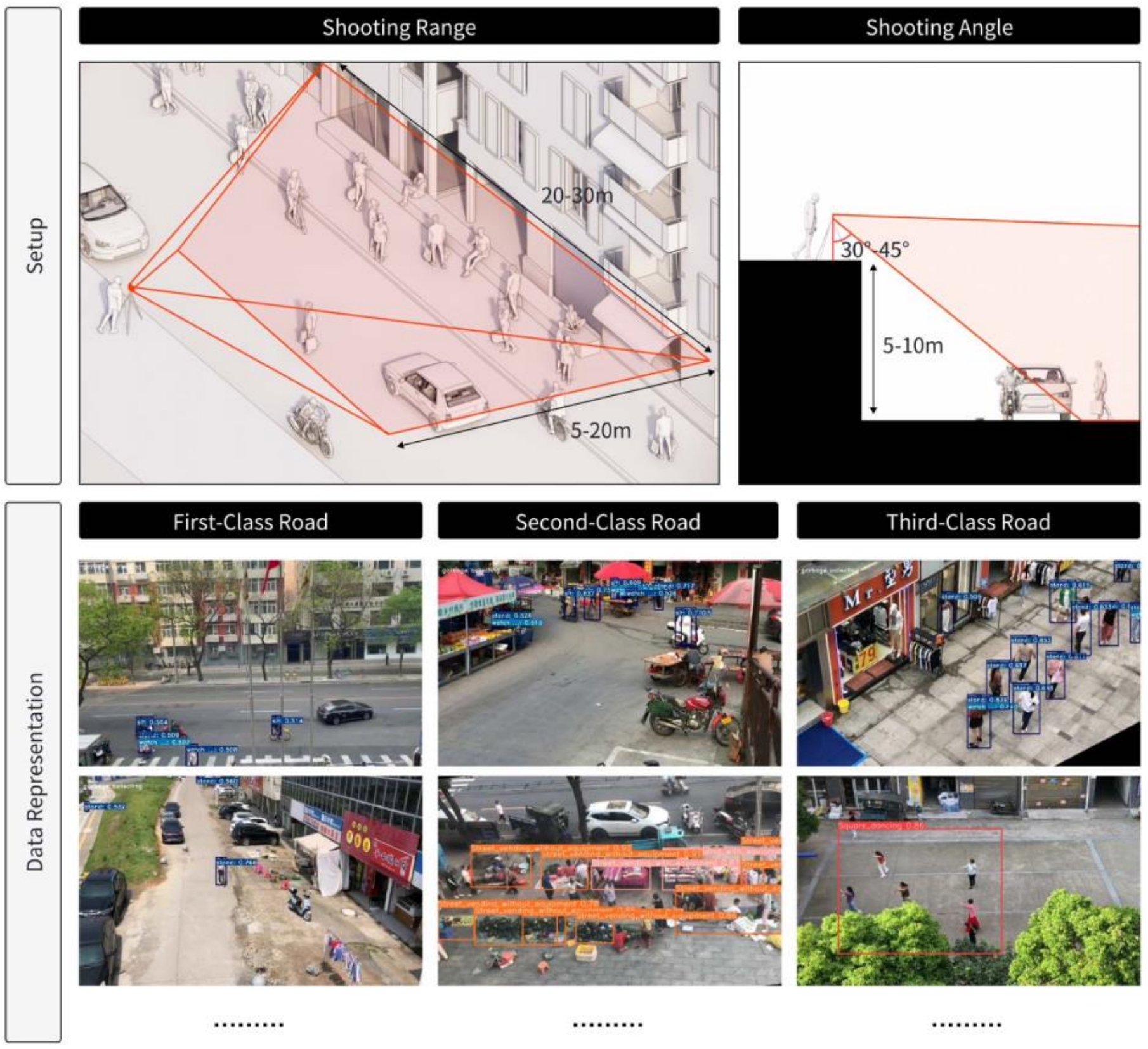

*Figure 2.* Data collection.

## 2.5 BEHAVIOR EVAL MODEL（BEM)

We employed a comprehensive approach to train and optimize a random forest regression model, namely BEM. First, the mmaction2 and YLX in BCN were used to infer and recognize our cw_set dataset. The combination of feature parameters identified in BCN was used as the input parameter "bem_set" for BEM, along with the corresponding street level labels for that set of feature parameters, to train the model's targeted predictive ability for different levels of streets. The final output is the score for the degree of informal behavior on the street.

### *2.5.1 Preprocessing*

We used PCA (Principal Component Analysis)(Pearson, 1901) to reduce the dimensionality of low-frequency or insignificant features in the cw_set dataset, determining the optimal dimensions through the explained variance ratio plot. This process retains enough information while reducing the number of features. The dataset was split into a training set and a test set at an 8:2 ratio, with behavior features not present in the inferred videos uniformly set to 0.

### *2.5.2 Training and Optimization*

Subsequently, we used the random forest algorithm to build the BEM, improving prediction accuracy through the integration of multiple decision trees. Each tree independently provided a prediction, and the final prediction was obtained by aggregating all tree predictions, typically determined by majority voting or averaging. Then, we optimized the model parameters, such as the number of trees, maximum number of features, and depth, through random search and RandomizedSearchCV to find the best configuration, reduce the risk of overfitting, and improve model accuracy.

### *2.5.3 Model Management*

We utilized the mlflow framework, released by Databricks at https://github.com/mlflow/mlflow, to manage the experimental process, including parameter selection, model training, and evaluation. Through mlflow, we could track experiment progress, compare the performance of different parameter combinations, and determine the optimal configuration. By combining random search with mlflow, experimental efficiency and reproducibility were greatly enhanced. During training and evaluation, we used matplotlib to visualize the explained variance ratio from PCA, sklearn to assess model performance, and joblib to save the final model for later deployment.

### 2.6 DATA ANNOTATION

We employed the X-AnyLabeling tool (https://github.com/CVHub520/X-AnyLabeling) to assist with the annotation tasks for the YLX model. This tool offers both automated and manual annotation features, significantly enhancing the efficiency and accuracy of the annotation process. The data annotation workflow commenced with the initial screening of the YLX dataset to identify high-quality samples—those that are clear, well-defined, and free from obstructive interference. Subsequently, we utilized X-AnyLabeling for automated annotation, followed by meticulous manual verification to ensure the precision and consistency of each annotation.

## 3. Result

### 3.1 MODEL EVALUATION

In the BCN model, the accuracy of the YLX was improved to 0.794 during the training phase by adjusting the learning rate and the number of learning iterations. The BEM's Mean Squared Error (MSE) was 0.9599, Root Mean Squared Error (RMSE) was 0.9798, and the Coefficient of Determination (R^2) was -0.1681.

We re-shot 10 videos, each 10 seconds long, according to the standard procedure to verify our model's performance on unseen data, i.e., to assess whether the model can be broadly applied. First, we submitted the videos to CommuniWave for inference to obtain predictions of the degree of informal behavior in the test samples. Then, we once again recruited 10 volunteers to score the videos according to the standard procedure, resulting in an average score for each video. The Mean Absolute Deviation (MAD) between CommuniWave's predictions and the volunteers' average scores was 0.709, indicating some deviation but overall close alignment with human ratings.

### 3.2 FEATURE IMPORTANCE

We used SHAP values (SHapley Additive exPlanations) to elucidate the key features contributing to the model's output. SHAP is a method derived from cooperative game theory that can visualize the impact and importance of individual features on the output (see figure 3). Across the three different street levels we set, informal behaviors showed varying degrees of influence. Street vending, both with and without equipment, had a significant impact on the degree of informal behavior on several streets; three-wheeled motorcycling and walking on the road particularly affected the order of first-class roads; gatherings of pedestrians on non-motorized lanes were the main influencing factor, along with square dancing, which also impacted the order of second-class roads; for third-class roads, informal behaviors predominantly involved street vending.

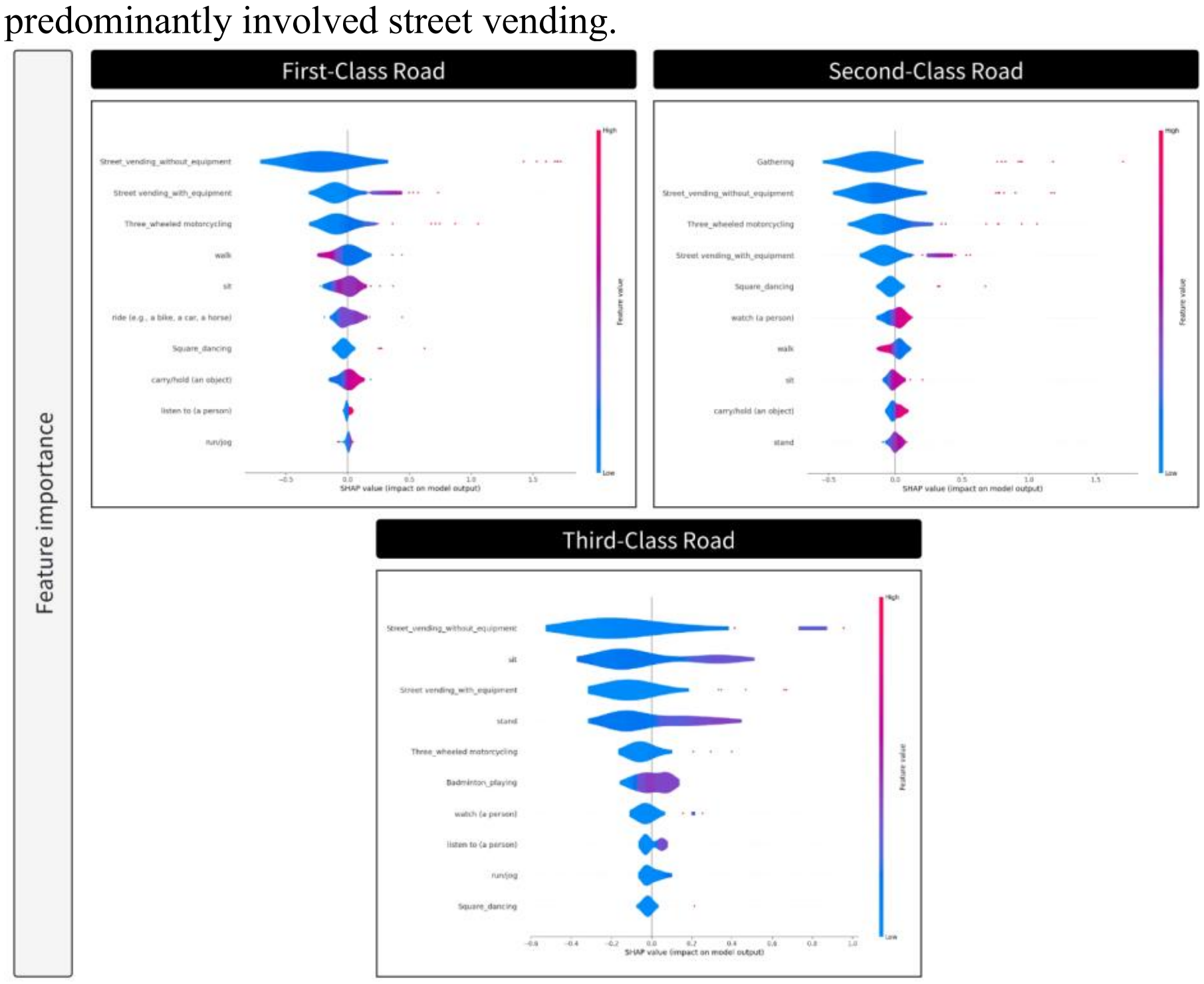

*Figure 3. Feature importance.*

### 3.3 DEGREE OF INFORMAL BEHAVIOR

To validate the practicality of our model, we recorded 10-minute videos during the morning, noon, and evening on representative streets from three different classes. Each video was segmented into 10-second clips and input into the CommuniWave model, which generated fluctuation charts displaying 60 predicted values of the Degree of Informal Behavior (DIB) (see Figure 4). The results indicate that on first-class roads, due to spontaneous activities such as street vending and pedestrian gatherings in the morning, DIB values remained high (between 4 and 5 points) before declining at noon and in the evening. A similar pattern was observed on second-class roads. Although there were occasional vending activities on third-class roads during the morning and noon, the DIB values remained relatively low (between 2 and 5 points).

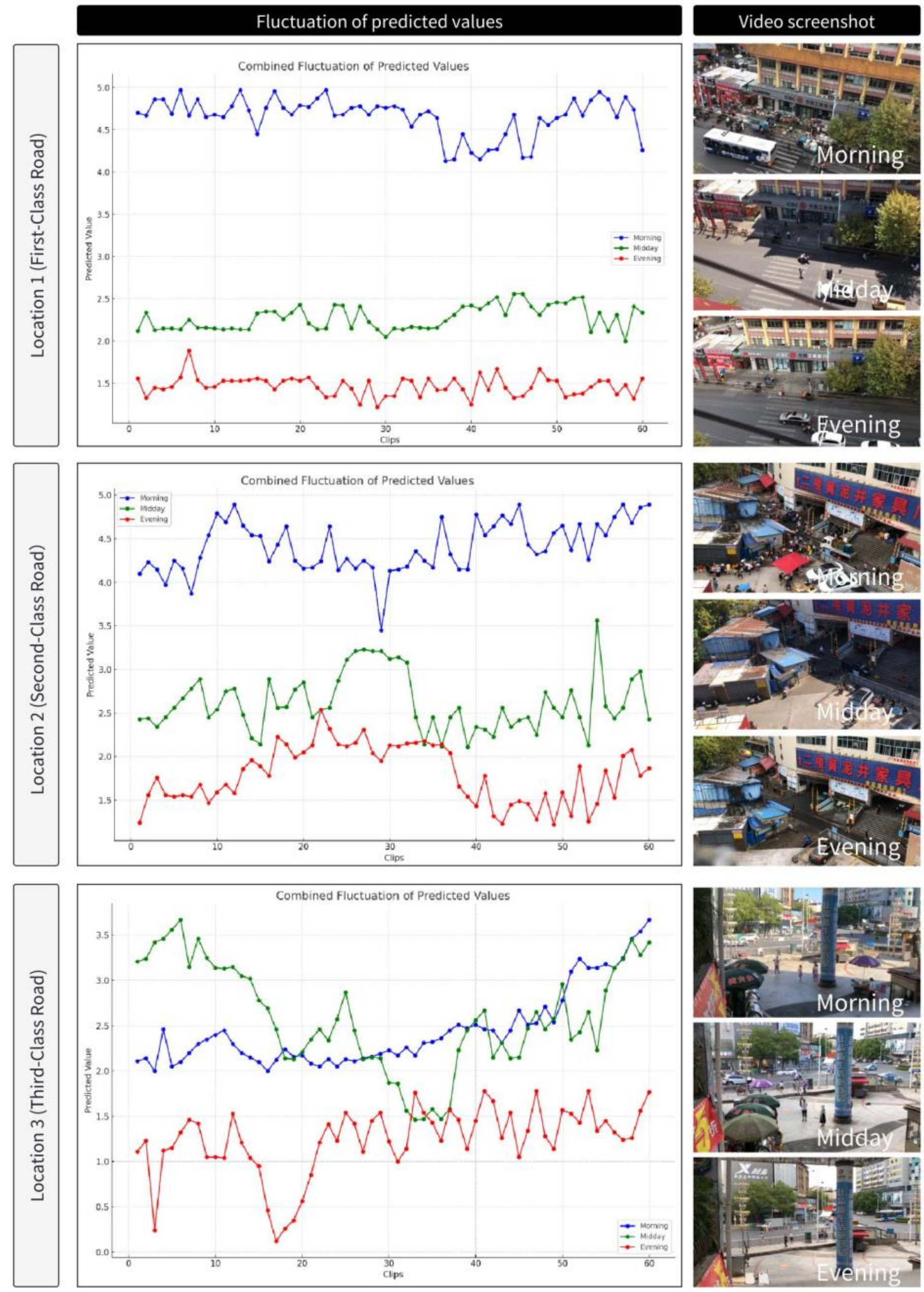

*Figure 4. The fluctuations of the DIB predicted by CommuniWave for three different road classes across three different time periods..*

## 4.Discussion

### 4.1 SUMMARY OF RESULTS

The feature importance analysis indicates that different types of informal behaviors have varying impacts on urban environments. As expected, informal street vending by residents is one of the primary factors contributing to higher DIB values in different urban neighborhoods. This finding aligns with previous studies, which highlight street vendors as a significant component of the informal urban economy(Onodugo et al., 2016). However, while vending activities play a prominent role in the DIB across different types of streets, each street class exhibits unique characteristics. For example, the driving and parking of tricycles on motorways, as well as pedestrians crossing roads, disrupt traffic order and are notable features on first-class roads, which aligns with urban traffic safety regulations. Meanwhile, activities involving crowd gatherings, such as square dancing, are prominent in the feature importance of second-class roads, reflecting residents' preference for spontaneous gathering spaces(Xiao, Hui and Wen, 2020). Notably, residents show increased tolerance for activities like square dancing on third-class roads, further revealing the close relationship between the standards by which people evaluate informal behaviors and the varying urban environments.

### 4.2 LIMITATIONS

Due to the randomness and unpredictability of informal behaviors in urban communities, BCN cannot fully identify all categories of informal behaviors. This may result in anomalies during video scoring tasks. It means that for different regions or street environments with varying socio-cultural contexts, BCN's label library cannot achieve complete coverage, necessitating the collection of customized datasets for secondary training. Additionally, although BCN performs well in street informal behavior recognition tasks, it typically only identifies surface features in images (videos) and lacks the ability to understand the deeper semantic content of the images (for example, the model cannot distinguish between a car parked on a sidewalk and one parked in a designated parking space). These issues may cause BEM to miss certain critical behavioral features in rare cases, leading to incorrect predictions of the degree of informal behavior. Moreover, since the mmaction and YOLOv10 models cannot perfectly handle videos in any input format, this may cause the program to crash. However, researchers can review the data output by BCN, and if any discrepancies are found, manual corrections can be applied to reduce the impact of recognition errors from BCN.(Das et al., 2019)

### 4.3 Conclusions

The main challenges in managing informal behaviors in urban communities include defining spontaneous behaviors and assessing them flexibly (Kapsalis and Kapsalis, 2020). CommuniWave effectively addresses both issues by providing urban designers with a machine learning framework that predicts the DIB based on street video data. Leveraging the powerful spatiotemporal action detection capabilities of mmaction2 and YOLOv10, the BCN within the model quickly identifies various spontaneous behaviors in the community. Compared to traditional field research methods, this significantly improves statistical efficiency.

This allows urban management departments to flexibly manage complex spontaneous behaviors in urban communities, avoiding conflicts and disputes caused by heavy-handed management approaches and enhancing the region's territorial resilience. Additionally, the model's dynamic DIB score intuitively reflects behavioral fluctuations within the location during the captured time period. Urban managers can implement different measures according to the DIB, addressing past fairness issues in urban management, where informal behaviors were often over- or under-regulated (Lane and McDonald, 2005). Notably, permitting informal behaviors within designated areas meets the daily needs of ordinary residents. This approach not only satisfies the top-down management requirements but also increases residents' participation and happiness (Dias et al., 2018), facilitating a sustainable urban community environment.

Furthermore, the model is adaptable. Urban management departments can pre-train the model based on the data collection methods (see 2.1 Data Collection) to suit the spontaneous behaviors in different urban and community environments. As a result, the model enhances the ability to regulate informal behaviors without undermining the vitality of spontaneous community actions, providing robust data support and decision-making guidance for creating more adaptive and resilient urban communities. This data-driven approach not only improves management flexibility but also mitigates the inequities caused by over- or under-management, helping to strengthen the overall resilience of communities in the face of future uncertainties. By integrating top-down planning requirements with bottom-up resident needs, the model lays the groundwork for sustainable urban development.